\documentclass{article}

\PassOptionsToPackage{numbers,compress}{natbib}

\usepackage[preprint]{neurips_2026}

\usepackage[utf8]{inputenc} 
\usepackage[T1]{fontenc}    
\usepackage{hyperref}       
\usepackage{url}            
\usepackage{booktabs}       
\usepackage{amsfonts}       
\usepackage{nicefrac}       
\usepackage{microtype}      

\usepackage[table]{xcolor}  
\usepackage{wrapfig}
\usepackage{multirow}
\usepackage{graphicx}
\usepackage{amsmath}
\usepackage{enumitem}
\usepackage{adjustbox}

\definecolor{groupbg}{RGB}{240,244,255}

\usepackage{xspace}
\newcommand{\model}{\textsc{LAC}\xspace}

\title{Latent Action Control for Reasoning-Guided Unified Image Generation}

\author{%
  \normalfont
  Fuxiang Zhai \quad
  Sixiang Chen \quad
  Yingjin Li \quad
  Shuaibo Li \\
  Jianyu Lai \quad
  Tengjun Huang \quad
  Lei Zhu \\
  The Hong Kong University of Science and Technology (Guangzhou) \\
  \texttt{fzhai2002@gmail.com}
}

\begin{document}

\maketitle

\begin{abstract}
Unified multimodal models can encode visual understanding and image generation within a shared backbone, yet understanding does not automatically translate into control: models may infer objects, relations, or knowledge cues but fail to instantiate them in the generated image. We propose \textbf{Latent Action Control (LAC)}, which makes reasoning actionable by representing it as hidden continuous actions inside a unified generator. Given a prompt, LAC rolls out a role-structured latent trajectory for planning, internal visual drafting, diagnosis, and refinement, and injects these actions into the hidden stream that conditions flow-based generation, without producing reasoning tokens or intermediate images. Since such action trajectories are unobserved, LAC learns them through prior-guided variational latent action alignment from training-only rendered semantic priors, draft image features, and supervised halting signals, followed by Latent-Flow GRPO to align the latent-to-image rollout with terminal visual feedback. This provides a control path from inferred relations, bindings, and knowledge cues to the generation process. Instantiated on BAGEL-7B-MoT, LAC consistently improves compositional and knowledge-grounded generation across GenEval, WISE, and T2I-CompBench, with the largest gains on spatial relations, attribute binding, and world-knowledge-sensitive prompts. Ablations and latent interventions show that the learned action trajectory is consumed by the generator, suggesting that unified generation benefits when understanding is not only encoded, but made actionable during generation.
\end{abstract}

\section{Introduction}

Unified multimodal models have substantially narrowed the architectural divide between visual understanding and image generation by placing both capabilities within a shared backbone. Recent systems~\cite{team2024chameleon,zhou2024transfusion,xie2024show,deng2025emerging,wu2025janus,wang2024emu3} show that unified architectures can support perception and synthesis within a common modeling framework. Yet sharing a backbone does not ensure that what the model infers during understanding becomes an effective control signal for generation. Recent studies reveal a persistent understanding--generation gap in unified models~\cite{niu2025does,lyu2025understanding}: a model may correctly infer objects, relations, or constraints from the input, but still fail to instantiate them faithfully in the generated image. This gap is most visible in prompts requiring compositional binding, spatial coordination, implicit world knowledge, or correction of likely generation errors. These observations point to a missing control interface: inferred constraints must be converted into generation-time conditioning signals before visual synthesis begins.

Existing approaches bridge the understanding--generation gap in three main ways. 
First, reasoning can be externalized as natural-language plans, guidance chains, visual checklists, or iterative edit instructions~\cite{liao2025imagegen,li2025visual,guo2025thinking,li2025editthinker,lin2025jarvisevo}. 
While such traces improve controllability and prompt fidelity, they remain textual or symbolic artifacts outside the generator's hidden-state dynamics, and must be reinterpreted as conditioning signals for synthesis. 
Second, auxiliary generation objectives or self-generated supervision can strengthen understanding--generation transfer~\cite{su2026generation,han2026unicorn}, but they do not specify how intermediate understanding states become executable generation-time control. 
Third, latent reasoning methods move deliberation into continuous representation spaces~\cite{li2025latent,wang2026regular,wang2025monet,mi2025milr,chen2026show}. 
However, these latent states are typically not formulated as role-specific actions written back into a unified generator to steer synthesis, nor are they optimized with the downstream flow-generation trajectory under terminal visual feedback. 
The remaining gap is therefore not simply where reasoning is represented, but whether it becomes an internal action space for controlling generation.

We propose \emph{Latent Action Control} (LAC), a framework that turns reasoning into an internal control interface for unified image generation. Before synthesis, LAC carries out a hidden-space latent deliberation process, where a latent policy head emits continuous latent actions and writes them back into the model's hidden stream. Given an input prompt, the resulting action trajectory is organized into four roles: \texttt{plan}, \texttt{draft}, \texttt{diagnosis}, and \texttt{refine}, corresponding to global intent planning, internal visual hypothesis formation, failure diagnosis, and targeted refinement. The \texttt{draft} role forms an internal visual hypothesis without emitting an intermediate image, while all latent actions directly condition the subsequent synthesis process. Rather than exposing reasoning as text or delaying it to post-hoc feedback, LAC makes understanding-side computation actionable during generation.

Because ground-truth action trajectories are unavailable, \model trains the hidden action interface with prior-guided variational latent action alignment. Instead of imitating decoded reasoning traces, it derives continuous action priors from training-only teacher signals. For \texttt{plan}, \texttt{diagnosis}, and \texttt{refine}, the priors come from structured teacher records rendered as compact document-style images; for \texttt{draft}, they come from visual features of a training-time imperfect candidate image. The resulting targets are removed at inference, but during training they align an executable autoregressive action policy, with supervised halting labels specifying role-wise computation lengths.

After supervised alignment, \model optimizes the full latent-to-image rollout with outcome-level feedback. We introduce \emph{Latent-Flow GRPO} (LF-GRPO), which assigns terminal image rewards to both continuous latent actions and flow-generation transitions through group-relative advantages. This lets image-level feedback shape the hidden actions that condition synthesis, rather than only the synthesis trajectory. Empirically, \model improves compositional and knowledge-grounded generation, especially on spatial relations, attribute binding, and world-knowledge grounding.

Our contributions are:
\begin{itemize}
    \item We formulate reasoning in unified image generation as \emph{Latent Action Control}, where role-specific continuous actions convert inferred constraints into generation-time conditioning.

    \item We introduce prior-guided variational latent action alignment and construct LAC-15K, a role-structured training set that provides rendered semantic priors, visual draft targets, and supervised halting signals for learning executable latent actions.

    \item We develop LF-GRPO, an outcome-level objective that jointly optimizes hidden actions and flow-generation transitions under shared terminal image rewards.
\end{itemize}
\section{Related Work}

\subsection{Unified Multimodal Models}

Unified multimodal models have increasingly reduced the architectural separation between visual understanding and image generation. Shared-token models~\cite{team2024chameleon,wang2024emu3} represent text and visual content in a common token space, while Janus~\cite{wu2025janus} retains a unified framework but decouples the visual pathways for understanding and generation. Hybrid systems~\cite{xie2024show,zhou2024transfusion} combine language modeling with diffusion-style synthesis, and recent unified models~\cite{deng2025emerging,tian2026internvl} further extend this paradigm toward understanding, reasoning, generation, and editing. Together, these systems show that perception and synthesis can coexist within a shared architecture. However, a shared architecture does not by itself specify how inferred constraints should become controllable generation-time states.

Recent work studies this understanding--generation gap more directly. Niu et al.~\cite{niu2025does} examine whether visual understanding transfers to generation, while UiG~\cite{lyu2025understanding} inject understanding signals into the synthesis process. Other approaches improve transfer through auxiliary generation objectives or self-generated supervision~\cite{su2026generation,han2026unicorn}. These methods strengthen cross-modal transfer through injected signals, additional objectives, or extra supervision. In contrast, \model places the bridge inside the unified generator: a latent policy emits role-structured continuous actions from hidden states and writes them back as generation-time control signals for image synthesis.

\subsection{Reasoning for Visual Generation}

Recent work uses intermediate reasoning to improve compositionality and prompt fidelity in visual generation. Reasoning is expressed through natural-language chains, stage-aware guidance, interleaved textual deliberation, visual reasoning traces, or critique-based edit instructions~\cite{liao2025imagegen,li2025visual,guo2025thinking,ye2025visual,li2025editthinker,lin2025jarvisevo}. Although these methods expose useful intermediate decisions, their reasoning is
typically externalized as text, checklists, traces, or edit instructions, rather
than embedded in the generator's native hidden-state dynamics.

A complementary line moves reasoning into continuous latent spaces. Prior work introduces latent reasoning states in language models~\cite{hao2022training,zhang2025soft} and extends them to multimodal perception and VQA~\cite{li2025latent,wang2026regular,wu2026lavit,zhang2026multimodal}. For visual generation, recent methods search over multimodal latents at test time, learn latent visual thoughts, or incorporate implicit latent reasoning into text-to-image generation~\cite{mi2025milr,wang2025monet,chen2026show}. These works show the promise of continuous reasoning, but typically treat latent states as representations to be searched, aligned, or consumed, rather than as executable actions written back into a unified generator to control synthesis. In contrast, \model formulates latent variables as role-structured control actions: the action trajectory is organized around planning, internal visual drafting, diagnosis, and refinement, and is trained with rendered semantic priors, visual draft targets, and downstream image-level feedback.

\subsection{Reinforcement Learning for Generation}

Reinforcement learning has become an effective tool for improving generative models beyond likelihood-based training. Learned reward models provide preference or alignment signals for text-to-image and multimodal generation~\cite{xu2023imagereward,kirstain2023pick,unifiedreward-think}. Policy- and preference-based methods have been applied to diffusion models~\cite{black2023training,fan2023dpok}, and recent online RL methods further extend to diffusion and flow-based visual generators~\cite{liu2025flow,xue2025dancegrpo}. These methods improve visual synthesis trajectories, but their policy variables mainly lie within the synthesis process rather than in intermediate reasoning states.

UniGRPO~\cite{liu2026unigrpo} is closest to our outcome-aligned training, as it jointly optimizes textual reasoning and flow-based image synthesis with terminal rewards. The key distinction is the policy space. UniGRPO optimizes explicit textual reasoning together with flow synthesis, whereas \model keeps deliberation in the hidden continuous space. LF-GRPO optimizes continuous latent actions and flow-generation transitions as a coupled rollout policy, allowing terminal image rewards to assign credit to both the hidden action trajectory and visual synthesis.
\section{Method}
\label{sec:method}

\subsection{Role-Structured Latent Action Control}
\label{sec:latent-action-control}

We consider a unified multimodal generator where text tokens, visual tokens, and image latents interact through a shared hidden stream. Instead of decoding reasoning into text, prompt rewrites, or edit instructions, \model turns reasoning into continuous latent actions that are written into this stream to control image synthesis. We instantiate \model on BAGEL~\cite{deng2025emerging}, while the formulation applies to any unified backbone whose hidden states can guide generation.

\begin{figure}[t]
  \centering
  \includegraphics[width=0.99\textwidth]{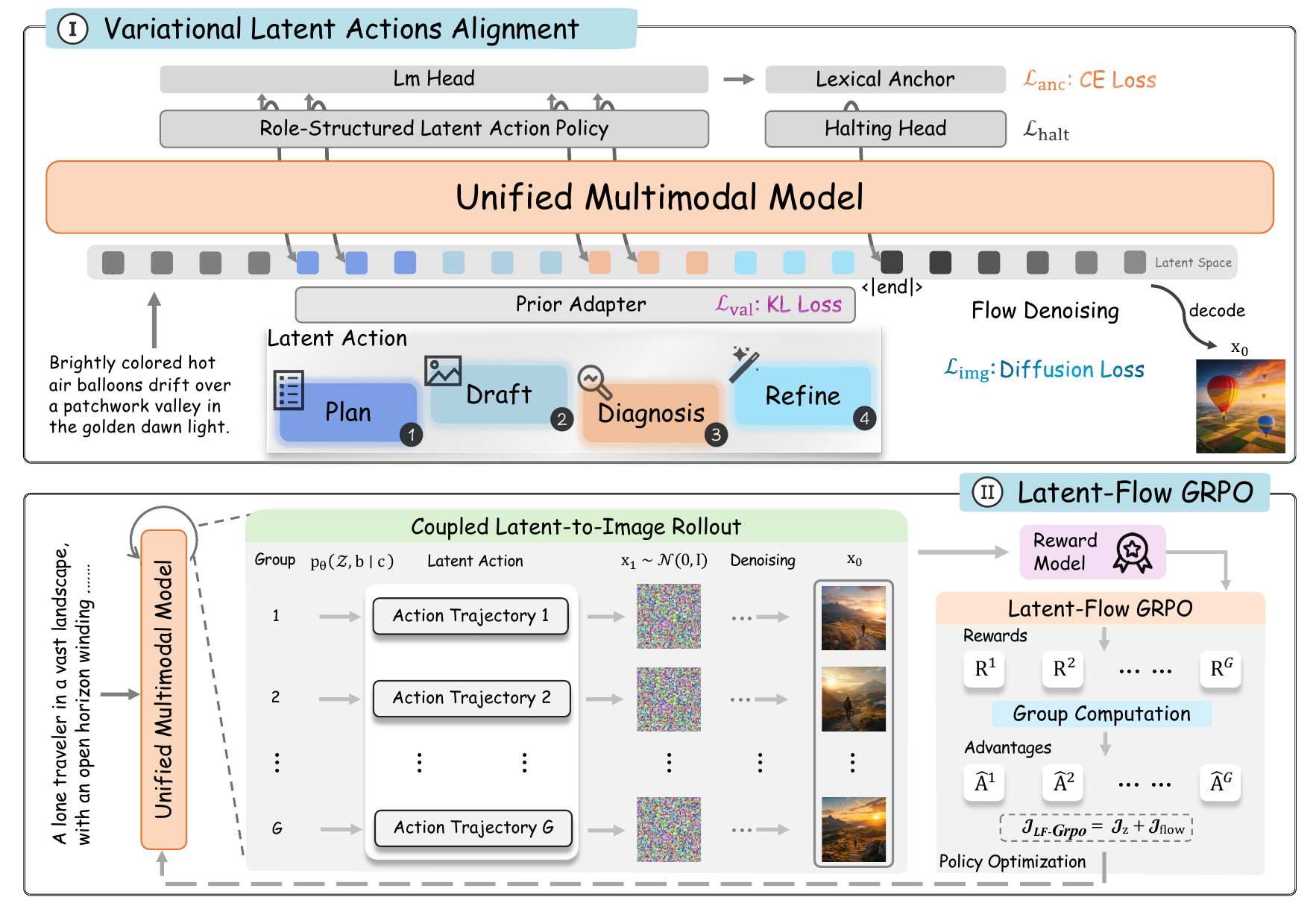}
  \caption{
  Overview of \model. The model rolls out a role-structured latent action trajectory with four hidden roles, conditions image synthesis on the resulting hidden states, and trains the hidden action interface with prior-guided latent action alignment followed by outcome-level Latent-Flow GRPO.
  }
  \label{fig:pipeline}
\end{figure}

Given a prompt $c$, \model samples a latent action trajectory organized into four ordered roles:
\texttt{plan}, \texttt{draft}, \texttt{diagnosis}, and \texttt{refine},
corresponding to planning, internal visual drafting, failure diagnosis, and
targeted refinement:
\begin{equation}
\label{eq:latent-sequence}
\mathcal Z =
\Big(
z^{\texttt{plan}}_{1:L_{\texttt{plan}}},
z^{\texttt{draft}}_{1},
z^{\texttt{diagnosis}}_{1:L_{\texttt{diagnosis}}},
z^{\texttt{refine}}_{1:L_{\texttt{refine}}}
\Big),
\qquad
z^s_k\in\mathbb R^d .
\end{equation}
Each $z^s_k$ is a continuous latent action. The \texttt{draft} role is single-step and forms an internal visual hypothesis, while the other roles have variable lengths bounded by $(L^s_{\min},L^s_{\max})$. During supervised alignment, teacher length labels supervise role-wise halting; at inference, a halting head stops the latent rollout and inserts a single \texttt{<end>} sentinel. We separate these components explicitly: $\mathcal Z$ contains only continuous actions, the halting decisions determine the realized length $T$, and \texttt{<end>} marks termination but is excluded from $\mathcal Z$ and the latent action head.

Let $H_\theta(c,\mathcal Z,\texttt{<end>})$ denote the conditioning hidden states
obtained after prompt encoding, latent action injection, and sentinel insertion.
The rollout distribution factorizes as
\begin{equation}
\label{eq:rollout-factorization}
\mathbb P_\theta(x_0,\mathcal Z,T\mid c)
=
p_\theta(\mathcal Z,T\mid c)\,
\pi_\theta\!\left(
x_0 \mid H_\theta(c,\mathcal Z,\texttt{<end>})
\right).
\end{equation}
In Eq.~\eqref{eq:rollout-factorization}, $p_\theta$ denotes the role-structured
latent action policy, $T$ is the realized latent length determined by halting,
and $\pi_\theta$ denotes the image generation policy conditioned on the resulting
hidden states. Both policies are implemented within the same backbone and
communicate through the hidden stream, enabling latent action control without
introducing explicit reasoning tokens.

\subsection{Prior-Guided Variational Latent Action Alignment}
\label{sec:variational-alignment}

The latent action trajectory is unobserved in training data. Direct textual trace supervision would reduce hidden actions to token imitation and may discard record fields, symbolic cues, and spatial layout. We therefore construct a teacher-induced latent prior used only during training. Following the rendered-supervision principle of~\cite{wang2026render}, we render structured teacher records into compact visual observations and encode them as dense target embeddings, which are removed at inference.

Given a prompt $c$, a reference image $\bar{x}$, and an imperfect candidate image
$\tilde{x}$, the teacher provides structured records $R^s_k$ for each semantic
role $s$ and step $k$. We construct the latent target $\eta^s_k$ as
\begin{equation}
\label{eq:latent-target-construction}
\eta^s_k
=
\begin{cases}
A_{\mathrm{prior}}
\!\left(
\mathrm{Pool}
\big(
E_{\mathrm{OCR}}(\mathrm{Render}(R^s_k))
\big)
\right),
&
s\in\{\texttt{plan},\texttt{diagnosis},\texttt{refine}\},
\\[2mm]
\alpha_{\mathrm{ld}}\,
\mathrm{LN}
\!\left(
\mathrm{Pool}_{\mathrm{nw}}(\tilde H)
\right),
&
s=\texttt{draft},\; k=1 .
\end{cases}
\end{equation}
Semantic-role targets are derived from rendered teacher records, while the
\texttt{draft} target is extracted from frozen visual features of $\tilde{x}$ as
a training-time prior. Here, $E_{\mathrm{OCR}}$ is frozen, $A_{\mathrm{prior}}$
is trainable, $\tilde H$ denotes frozen features of $\tilde{x}$, and
$\alpha_{\mathrm{ld}}$ rescales the draft target. Dataset details are given in
Appendix~\ref{app:dataset_construction}.

To disambiguate roles and steps, we augment each base target with identity
embeddings,
$e^s_k=\eta^s_k+e^{\mathrm{role}}_s+e^{\mathrm{step}}_k$.
Collecting these vectors under the predefined role schedule gives
$E=\{e_n\}_{n=1}^{N}$, a sequence of prior means in one-to-one correspondence
with latent actions $\mathcal Z=\{z_n\}_{n=1}^{N}$. Rather than using $E$ as
point supervision, we define a fixed-variance Gaussian prior over latent actions:
\begin{equation}
\label{eq:latent-action-prior}
p_\gamma(\mathcal Z\mid\mathcal U)
=
\prod_{n=1}^{N}
\mathcal N
\!\left(
z_n; e_n, \sigma_q^2 I
\right),
\end{equation}
where $\mathcal U$ denotes the teacher information used to construct $E$, and
$\sigma_q$ is fixed. In Eq.~\eqref{eq:latent-action-prior}, the
\texttt{<end>} sentinel is excluded because it marks a deterministic boundary
rather than a latent action.

We train the latent policy as an executable autoregressive rollout rather than by teacher forcing. At step $n$, $h_n$ is computed from the prompt, the sampled action history $z_{\le n}$, and the scheduled role boundaries. The policy predicts
\begin{equation}
\label{eq:latent-policy}
q_\theta(z_{n+1}\mid h_n)
=
\mathcal N
\!\left(
\mu_{n+1}(h_n),
\operatorname{diag}(\sigma^2_{n+1}(h_n))
\right),
\end{equation}
samples $z_{n+1}$ by reparameterization, and writes it back into the hidden stream. The target $e_{n+1}$ is never injected as input; it only defines the local teacher prior $\mathcal N(e_{n+1},\sigma_q^2 I)$. Thus, the policy is conditioned on the executable hidden state induced by its own sampled rollout.

The variational alignment loss is
\begin{equation}
\label{eq:variational-alignment-loss}
\mathcal L_{\mathrm{var}}
=
\mathbb E_{\epsilon_{1:N-1}}
\left[
\sum_{n=0}^{N-1}
D_{\mathrm{KL}}
\left(
\mathcal N(e_{n+1},\sigma_q^2 I)
\,\|\, 
q_\theta(\cdot\mid h_n)
\right)
\right].
\end{equation}
This aligns the policy under its own sampled rollout while keeping teacher signals as training-time priors. We combine this objective with a lexical anchor loss $\mathcal L_{\mathrm{anc}}$ that weakly grounds latent states in teacher semantics, a supervised halting loss $\mathcal L_{\mathrm{halt}}$, and an image flow-matching loss $\mathcal L_{\mathrm{img}}$ conditioned on $H_\theta(c,\mathcal Z,\texttt{<end>})$. Details are provided in Appendix~\ref{app:method_losses}. The supervised fine-tuning objective is
\begin{equation}
\label{eq:sft-objective}
\mathcal L_{\mathrm{SFT}}
=
\lambda_{\mathrm{var}}\mathcal L_{\mathrm{var}}
+
\lambda_{\mathrm{anc}}\mathcal L_{\mathrm{anc}}
+
\lambda_{\mathrm{halt}}\mathcal L_{\mathrm{halt}}
+
\lambda_{\mathrm{img}}\mathcal L_{\mathrm{img}}.
\end{equation}

Supervised fine-tuning has two stages. The correction warm-up stage uses $(c,\tilde{x})$ to train only \texttt{diagnosis} and \texttt{refine}, with $\tilde{x}$ encoded through the visual perception path and $\lambda_{\mathrm{img}}=0$. The full generation stage uses only $c$, activates all four roles, and jointly trains latent actions with image synthesis. Although the \texttt{draft} target comes from a training-time imperfect candidate, no intermediate image is emitted or provided at inference. At inference, all teacher-derived signals are removed: the model receives only $c$, samples $\mathcal Z$, halts with the halting head, inserts a single \texttt{<end>} sentinel, and synthesizes the image from the resulting hidden states.

\subsection{Outcome-Aligned Latent-Flow GRPO}
\label{sec:lf-grpo}

Prior-guided alignment initializes the latent action policy, but teacher priors optimize only local action targets and do not directly optimize final image quality. Inspired by UniGRPO~\cite{liu2026unigrpo}, we introduce \emph{Latent-Flow GRPO} (LF-GRPO), which optimizes continuous latent actions and flow-generation transitions under shared terminal rewards.

For each prompt $c$, we sample a group of $G$ rollouts
\begin{equation}
\label{eq:rollout-tuple}
\tau_i=(\mathcal Z_i,T_i,\mathcal X_i),
\qquad i=1,\ldots,G,
\end{equation}
where $\mathcal Z_i$ is the latent action trajectory, $T_i$ is the realized latent length determined by the halting head, and
$\mathcal X_i=\{x_{m,i}\}_{m=0}^{M}$ is the flow-generation trajectory, with
$x_{0,i}$ denoting the final image. Let
$H_i=H_\theta(c,\mathcal Z_i,\texttt{<end>})$ be the conditioning hidden states
after the latent rollout terminates. The halting decision selects the active
latent positions, but its Bernoulli log-probability is excluded from the RL
objective. Thus, LF-GRPO optimizes
\begin{equation}
\label{eq:lf-grpo-policy}
\Pi^{\mathrm{LF}}_\theta(\tau_i\mid c)
=
p^z_\theta(\mathcal Z_i\mid c,T_i)\,
\pi_\theta(\mathcal X_i\mid H_i),
\end{equation}
where $p^z_\theta$ is the product of Gaussian densities over active latent actions.

To reduce within-group variance, rollouts for the same prompt share the same
initial flow noise. Each rollout receives a terminal image reward
$R_i=R(c,x_{0,i})$, with reward model details in
Appendix~\ref{app:method_reward}. The group-relative advantage is
\begin{equation}
\label{eq:group-relative-advantage}
\hat A_i
=
\frac{
R_i-\operatorname{mean}(\{R_j\}_{j=1}^{G})
}{
\operatorname{std}(\{R_j\}_{j=1}^{G})
}.
\end{equation}
We assign the same $\hat A_i$ to all optimized latent actions and flow
transitions in $\tau_i$, giving terminal image feedback a credit path to both
the hidden action trajectory and visual synthesis.

\paragraph{Latent-action update.}
For each active latent action $z_{i,n}$, we compute the latent policy ratio
\begin{equation}
\label{eq:latent-policy-ratio}
\rho^z_{i,n}
=
\exp
\left(
\operatorname{clamp}
\left(
\ell^z_{\theta,i,n}-\ell^z_{\mathrm{old},i,n},
-\kappa,\kappa
\right)
\right),
\end{equation}
where $\ell^z_{\theta,i,n}$ and $\ell^z_{\mathrm{old},i,n}$ are normalized
Gaussian log probabilities under the current and old latent action policies.
The clamp stabilizes exponentiation, while the clipping below defines the GRPO surrogate.
Let $w^z_{i,n}$ mask active latent positions and
$W_z=\sum_{i,n}w^z_{i,n}$. The latent-action objective is
\begin{equation}
\label{eq:latent-action-grpo-loss}
\mathcal L_z
=
-\frac{1}{W_z}
\sum_{i=1}^{G}
\sum_{n=1}^{T_i}
w^z_{i,n}
\min
\left(
\rho^z_{i,n}\hat A_i,\,
\operatorname{clip}(\rho^z_{i,n},1-\varepsilon,1+\varepsilon)\hat A_i
\right)
+
\frac{\beta_z}{W_z}
\sum_{i=1}^{G}
\sum_{n=1}^{T_i}
w^z_{i,n}D^z_{i,n},
\end{equation}
where $D^z_{i,n}$ regularizes the latent action policy toward the reference policy.
Masking, log-probability normalization, and regularizer details are provided in
Appendix~\ref{app:method-la-grpo}.

\paragraph{Flow update.}
Following UniGRPO~\cite{liu2026unigrpo}, we optimize the image-side flow policy
on a subset of SDE timesteps $\Omega_{\mathrm{SDE}}$. Given conditioning hidden
states $H_i$, the clipped flow objective is
\begin{equation}
\label{eq:flow-objective}
\mathcal L_{\mathrm{flow}}
=
-\frac{1}{|\Omega_{\mathrm{SDE}}|G}
\sum_{m\in\Omega_{\mathrm{SDE}}}
\sum_{i=1}^{G}
\min
\left(
\tilde\rho^x_{i,m}\hat A_i,\,
\operatorname{clip}(\tilde\rho^x_{i,m},1-\varepsilon_x,1+\varepsilon_x)\hat A_i
\right),
\end{equation}
where $\tilde\rho^x_{i,m}$ is the behavior-corrected transition ratio of the
flow policy conditioned on $H_i$. To keep the actor close to the reference
generator, we regularize the predicted velocity field:
\begin{equation}
\label{eq:velocity-regularizer}
\mathcal R_{\mathrm{vel}}
=
\frac{1}{|\Omega_{\mathrm{SDE}}|G}
\sum_{m\in\Omega_{\mathrm{SDE}}}
\sum_{i=1}^{G}
\frac{1}{P_i d_x}
\sum_{u=1}^{P_i}
\left\|
v^{(u)}_{\theta}(x_{m,i},t_m,H_i)
-
v^{(u)}_{\mathrm{ref}}(x_{m,i},t_m,H_i)
\right\|_2^2 ,
\end{equation}
where $v_\theta$ and $v_{\mathrm{ref}}$ are the actor and reference velocities,
$P_i$ is the number of packed image tokens in rollout $i$, and $d_x$ is the
token dimension. Transition-ratio computation, token aggregation, and
stabilization details are provided in Appendix~\ref{app:method-flow-grpo}.

\paragraph{Overall objective.}
The final LF-GRPO objective is
\begin{equation}
\label{eq:lf-grpo-objective}
\mathcal L_{\mathrm{LF\text{-}GRPO}}
=
\lambda_z\mathcal L_z
+
\lambda_x\mathcal L_{\mathrm{flow}}
+
\beta_{\mathrm{vel}}\mathcal R_{\mathrm{vel}} .
\end{equation}
LF-GRPO jointly optimizes hidden latent actions and flow-based synthesis, rather
than updating only the image-generation trajectory. This encourages the
planning, drafting, diagnosis, and refinement actions to become aligned with
final image quality.
\section{Experiments}
\label{sec:experiments}

\subsection{Implementation Details}
\label{sec:experimental_setup}

\paragraph{Dataset.}
We construct \model-15K, a 15K-example training set for supervised latent action alignment and generation-coupled training. Each example starts from a prompt and pairs a high-quality target image from a stronger external generator with an imperfect draft image from the base model. From this pair, we derive structured teacher records for planning, diagnosis, and refinement. The semantic records are rendered into OCR-based prior features, while the imperfect draft is encoded into visual draft features. Additional construction details are provided in Appendix~\ref{app:dataset_construction}.

\paragraph{Training details.}
We instantiate \model on BAGEL-7B-MoT~\citep{deng2025emerging} and train it in three stages. Phase 1 performs correction-oriented supervised alignment for 2,500 steps, updating the understanding-side LoRA adapters and latent action modules while disabling image generation. Phase 2 starts from the step-2,000 checkpoint and performs generation-coupled alignment for 3,000 steps, training the full latent action trajectory to condition image synthesis. Phase 3 applies LF-GRPO for 250 updates with group size 8, using UnifiedReward-2.0~\citep{unifiedreward-think} served with vLLM to compute terminal image rewards and group-relative advantages. LF-GRPO updates the continuous latent action policy and flow-generation transitions while keeping the halting head fixed. All stages run on a single 8$\times$NVIDIA A800 node, with full details in Appendix~\ref{app:training_details}.

\subsection{Quantitative Evaluation}
\label{sec:quantitative}

\paragraph{GenEval.}
We evaluate compositional text-to-image generation on GenEval~\citep{ghosh2023geneval}. 
Under original prompts, \model obtains an overall score of 0.82, improving over our reproduced BAGEL baseline~\citep{deng2025emerging} by 0.05. 
Since object-presence categories are near saturation, the gains mainly appear in constraint-sensitive categories, including Colors (+0.13), Position (+0.11), and Color Attribute (+0.05), while Counting slightly decreases (-0.02). 
With LLM-rewritten prompts, \model$^{\dagger}$ improves over BAGEL$^{\dagger}$ from 0.86 to 0.90, with the largest gain again on Position (+0.12). 
These results suggest that \model improves the grounding of compositional constraints, especially attribute binding and spatial relations, rather than simply increasing object occurrence.

\begin{table}[t]
\centering
\caption{
Results on GenEval.
``Gen. Only'' and ``Unified'' denote image-only and unified multimodal generators, respectively.
$^{*}$: reproduced with official code/checkpoints; $^{\dagger}$: LLM-rewritten prompts.
}
\label{tab:geneval_result}
\begin{adjustbox}{width=\linewidth}
\begin{tabular}{lccccccc}
\toprule
\multicolumn{1}{c}{\bf Method} &
{\bf Single Obj. $\uparrow$} &
{\bf Two Obj. $\uparrow$} &
{\bf Counting $\uparrow$} &
{\bf Colors $\uparrow$} &
{\bf Position $\uparrow$} &
{\bf Color Attr. $\uparrow$} &
{\bf Overall $\uparrow$}
\\
\cmidrule{1-8}
\multicolumn{8}{c}{\textit{Generation Only}}\\
\cmidrule{1-8}
PixArt-$\Sigma$~(\citet{chen2024pixart}) & 0.98 & 0.50 & 0.44 & 0.80 & 0.08 & 0.07 & 0.48 \\
DALL-E 3~(\citet{betker2023improving}) & 0.96 & 0.87 & 0.47 & 0.83 & 0.43 & 0.45 & 0.67 \\
SD3-Medium~(\citet{esser2024scaling}) & 0.99 & 0.94 & 0.72 & 0.89 & 0.33 & 0.60 & 0.74 \\
FLUX.1-dev$^{\dagger}$~(\citet{flux2024}) & 0.98 & 0.93 & 0.75 & 0.93 & 0.68 & 0.65 & 0.82 \\
\cmidrule{1-8}
\multicolumn{8}{c}{\textit{Unified Multimodal}}\\
\cmidrule{1-8}
Chameleon~(\citet{team2024chameleon}) & {--} & {--} & {--} & {--} & {--} & {--} & 0.39 \\
SEED\hbox{-}X~(\citet{ge2024seed}) & 0.97 & 0.58 & 0.26 & 0.80 & 0.19 & 0.14 & 0.49 \\
TokenFlow\hbox{-}XL~(\citet{qu2025tokenflow})& 0.95 & 0.60 & 0.41 & 0.81 & 0.16 & 0.24 & 0.55 \\
Transfusion~(\citet{zhou2024transfusion}) & {--} & {--} & {--} & {--} & {--} & {--} & 0.63 \\
Emu3\hbox{-}Gen$^{\dagger}$~(\citet{wang2024emu3}) & 0.99 & 0.81 & 0.42 & 0.80 & 0.49 & 0.45 & 0.66 \\
Show\hbox{-}o2$^{\dagger}$~(\citet{xie2025show}) & 1.00 & 0.87 & 0.58 & 0.92 & 0.52 & 0.62 & 0.76 \\
MetaQuery-XL$^{\dagger}$~(\citet{pan2025transfer}) & {--} & {--} & {--} & {--} & {--} & {--} & 0.80 \\
Janus\hbox{-}Pro\hbox{-}7B~(\citet{chen2025janus}) & 0.99 & 0.89 & 0.59 & 0.90 & 0.79 & 0.66 & 0.80 \\
Mogao$^{\dagger}$~(\citet{liao2025mogao}) & 1.00 & 0.97 & 0.83 & 0.93 & 0.84 & 0.80 & 0.89 \\
\cmidrule{1-8}
BAGEL$^{*}$~(\citet{deng2025emerging}) & 0.99 & 0.94 & 0.77 & 0.84 & 0.50 & 0.60 & 0.77 \\  
\rowcolor{groupbg}
\textbf{\model(Ours)} & \textbf{1.00} & \textbf{0.95} & 0.75 & 0.97 & 0.61 & 0.65 & \textbf{0.82} \\
BAGEL$^{\dagger}$~(\citet{deng2025emerging}) & 0.99 & 0.94 & 0.80 & 0.95 & 0.70 & 0.78 & 0.86 \\
\rowcolor{groupbg}
\textbf{\model(Ours)$^{\dagger}$} & 0.99 & \textbf{0.95} & \textbf{0.83} & \textbf{0.99} & \textbf{0.82} & \textbf{0.81} & \textbf{0.90} \\
\bottomrule
\end{tabular}
\end{adjustbox}
\vspace{-0.6em}
\end{table}

\begin{table}[t]
\centering
\caption{
Results on WISE.
``Gen. Only'' and ``Unified'' denote image-only and unified multimodal generators, respectively.
Overall is computed with the official WISE weighted aggregation; BAGEL and our results are reported under the same protocol.
}
\label{tab:wise_result}
\begin{adjustbox}{width=\linewidth}
\begin{tabular}{lccccccc}
\toprule
\multicolumn{1}{c}{\bf Method} &
{\bf Culture $\uparrow$} &
{\bf Time $\uparrow$} &
{\bf Space $\uparrow$} &
{\bf Biology $\uparrow$} &
{\bf Physics $\uparrow$} &
{\bf Chemistry $\uparrow$} &
{\bf Overall $\uparrow$}
\\
\cmidrule{1-8}
\multicolumn{8}{c}{\textit{Generation Only}}\\
\cmidrule{1-8}
SD3.5-large~(\citet{esser2024scaling}) & 0.44 & 0.50 & 0.58 & 0.44 & 0.52 & 0.31 & 0.46 \\
PixArt-$\Sigma$~(\citet{chen2024pixart}) & 0.45 & 0.50 & 0.48 & 0.49 & 0.56 & 0.34 & 0.47 \\
Playground-v2.5 (\citet{li2024playground}) & 0.49 & 0.58 & 0.55 & 0.43 & 0.48 & 0.33 & 0.49 \\
FLUX.1-dev~(\citet{flux2024}) & 0.48 & 0.58 & 0.62 & 0.42 & 0.51 & 0.35 & 0.50\\
\cmidrule{1-8}
\multicolumn{8}{c}{\textit{Unified Multimodal}}\\
\cmidrule{1-8}
Janus~(\citet{wu2025janus}) & 0.16 & 0.26 & 0.35 & 0.28 & 0.30 & 0.14 & 0.23 \\
VILA-U~(\citet{wu2024vila}) & 0.26 & 0.33 & 0.37 & 0.35 & 0.39 & 0.23 & 0.31 \\
Show\hbox{-}o~(\citet{xie2024show}) & 0.28 & 0.40 & 0.48 & 0.30 & 0.46 & 0.30 & 0.35 \\
Janus\hbox{-}Pro\hbox{-}7B~(\citet{chen2025janus}) & 0.30 & 0.37 & 0.49 & 0.36 & 0.42 & 0.26 & 0.35 \\
Emu3\hbox{-}Gen~(\citet{wang2024emu3}) & 0.34 & 0.45 & 0.48 & 0.41 & 0.45 & 0.27 & 0.39 \\
Show\hbox{-}o2~(\citet{xie2025show}) & 0.33 &0.42 &0.53 &0.39   & 0.45 &0.26 & 0.39 \\
\cmidrule{1-8}
BAGEL$^{*}$~(\citet{deng2025emerging}) & 0.44 & 0.55 & 0.61 & 0.41 & 0.61 & 0.36 & 0.49 \\
\rowcolor{groupbg}
\textbf{\model(Ours)} & 0.49 & 0.57 & 0.65 & 0.59 & 0.60 & 0.43 & \textbf{0.54}\\
BAGEL w/ thinking$^{*}$~(\citet{deng2025emerging}) & \textbf{0.73} & 0.68 & 0.76 & 0.64 & 0.74 & 0.60 & 0.70 \\
\rowcolor{groupbg}
\textbf{\model w/ thinking (Ours)} & 0.71 & \textbf{0.76} & \textbf{0.77} & \textbf{0.69} & \textbf{0.78} & \textbf{0.65} & \textbf{0.73} \\
\bottomrule
\end{tabular}
\end{adjustbox}
\vspace{-0.6em}
\end{table}

\paragraph{WISE.}
Table~\ref{tab:wise_result} evaluates world-knowledge-informed generation on WISE~\citep{niu2025wise}. 
Using the official WISE weighted aggregation, \model improves over our reproduced BAGEL baseline from 0.49 to 0.54, driven by gains in Biology (+0.18), Culture (+0.05), and Chemistry (+0.07). 
With thinking, \model w/ thinking improves BAGEL w/ thinking from 0.70 to 0.73, improving Time, Biology, Physics, and Chemistry despite a slight decrease on Culture. 
Together with GenEval, these results suggest that \model improves both explicit compositional grounding and implicit world-knowledge grounding.

Overall, GenEval and WISE show a consistent pattern: \model yields the largest gains when generation requires grounding structured constraints beyond object presence, including spatial relations, attribute binding, temporal reasoning, and scientific knowledge.

\subsection{Analysis of Latent Reasoning for Generation}
\label{sec:latent-analysis}

\paragraph{Training pathway analysis.}
We isolate the effects of image supervision, latent capacity, latent action alignment, and outcome-level feedback using the same BAGEL~\citep{deng2025emerging} backbone and supervised prompt--image pairs. \emph{Img-SFT} uses only supervised flow training. \emph{Latent-SFT}$_{\mathrm{NR}}$ adds latent actions with non-rendered teacher targets. \emph{LAC-SFT} adds rendered semantic priors, visual draft grounding, and supervised halting. \emph{LAC} further applies LF-GRPO with terminal image rewards.

\begin{table}[tpbh]
\centering
\caption{Controlled training pathway analysis of latent action control, using LLM-rewritten prompts for GenEval$^\dagger$ and the official overall score for WISE.}
\label{tab:latent-analysis}
\scriptsize
\setlength{\tabcolsep}{2.6pt}
\renewcommand{\arraystretch}{1.08}
\begin{adjustbox}{width=\linewidth}
\begin{tabular}{lcccccccc}
\toprule
\textbf{Method} &
\multicolumn{7}{c}{\textbf{GenEval}$^\dagger$} &
\multicolumn{1}{c}{\textbf{WISE}} \\
\cmidrule(lr){2-8}
\cmidrule(lr){9-9}
&
\textbf{Single Obj.} $\uparrow$ &
\textbf{Two Obj.} $\uparrow$ &
\textbf{Counting} $\uparrow$ &
\textbf{Colors} $\uparrow$ &
\textbf{Position} $\uparrow$ &
\textbf{Color Attr.} $\uparrow$ &
\textbf{Overall} $\uparrow$ &
\textbf{Overall} $\uparrow$ \\
\midrule
BAGEL~\citep{deng2025emerging}
& 0.99 & 0.94 & 0.80 & 0.95 & 0.70 & 0.78 & 0.86 & 0.49 \\
Img-SFT
& 0.99 & 0.94 & 0.77 & \textbf{0.99} & 0.74 & 0.77 & 0.87 & 0.50 \\
Latent-SFT$_{\mathrm{NR}}$
& 0.98 & 0.95 & 0.75 & 0.97 & 0.75 & 0.79 & 0.87 & 0.50 \\
\model-SFT
& \textbf{1.00} & \textbf{0.96} & 0.81 & 0.98 & 0.79 & 0.78 & 0.89 & 0.52 \\
\model
& 0.99 & 0.95 & \textbf{0.83} & \textbf{0.99} & \textbf{0.82} & \textbf{0.81} & \textbf{0.90} & \textbf{0.54} \\
\bottomrule
\end{tabular}
\end{adjustbox}
\vspace{-0.6em}
\end{table}

Table~\ref{tab:latent-analysis} separates image supervision, latent capacity,
and alignment. Img-SFT gives only marginal gains over BAGEL (+0.01 on both
benchmarks), while Latent-SFT$_{\mathrm{NR}}$ remains at 0.87/0.50, showing
that neither extra image targets nor latent tokens alone explain the improvement.
The first clear gain comes from \model{}-SFT, which raises GenEval from 0.87 to
0.89 and WISE from 0.50 to 0.52, especially on Counting (+0.06) and Position
(+0.04). This suggests that rendered semantic targets and visual draft grounding
are crucial for turning latent actions into generation-useful hidden control.

LF-GRPO further improves \model-SFT to 0.90 on GenEval$^\dagger$ and 0.54 on WISE, mainly on Counting, Position, and Color Attribute. These categories require preserving object counts, spatial relations, and attribute bindings during synthesis, where image-level feedback can calibrate hidden actions toward the final outcome. The gain is consistent but smaller than that from supervised alignment, suggesting that LF-GRPO refines an already useful action space. It is also more compute-intensive and less stable than SFT; additional details are provided in Appendix~\ref{app:lfgrpo_challenges}.

\paragraph{Inference-time latent intervention.}

The training-pathway ablation identifies which objectives improve performance, but does not verify whether the sampled action trajectory is used during synthesis. We therefore intervene only on the latent actions of the same trained \model{} checkpoint, while keeping the prompt, model parameters, generation procedure, and latent budget fixed. \emph{Zero Latents} replaces active actions with zero vectors, \emph{Random Latents} uses random vectors with matched scale, and \emph{Shuffled Roles} permutes role-specific action blocks while preserving the number of actions.

\begin{figure}[!h]
\centering
\includegraphics[width=0.91\textwidth]{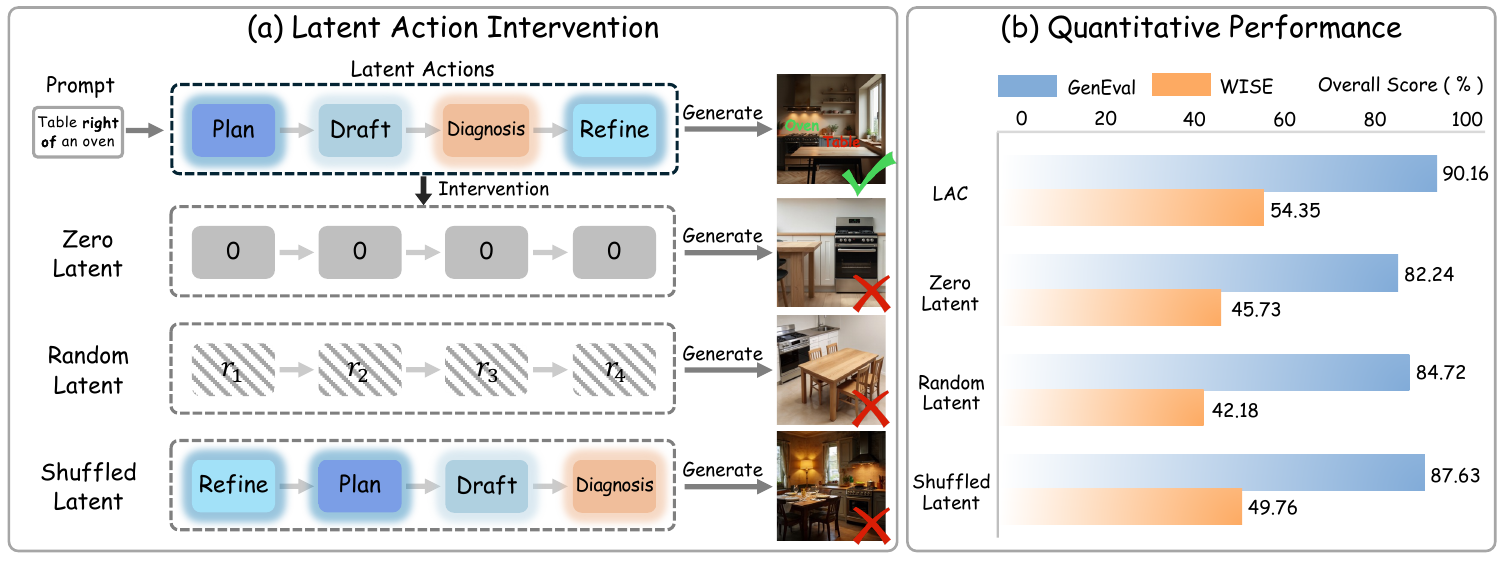}
\caption{
Inference-time latent intervention. Zero and random latents test dependence on action content, while role shuffling tests the functional order of latent actions.
}
\label{fig:latent-intervention}
\end{figure}

Figure~\ref{fig:latent-intervention} shows that the learned action trajectory is actively used during synthesis. Zeroing latent actions drops GenEval and WISE by 7.92 and 8.62 points, while replacing them with random actions drops the two scores by 5.44 and 12.17 points. The larger WISE degradation under random actions suggests that incorrect hidden actions can inject conflicting knowledge cues. Role shuffling causes a smaller but consistent drop of 2.53 and 4.59 points, showing that the ordered \texttt{plan}--\texttt{draft}--\texttt{diagnosis}--\texttt{refine} structure is also functional.

Together, Table~\ref{tab:latent-analysis} and Figure~\ref{fig:latent-intervention} rule out two simpler explanations. The training-pathway results show that the gains are not explained by extra image supervision or latent capacity alone, while the intervention results show that the learned actions are consumed rather than ignored at inference time. Thus, \model{} improves generation by making inferred constraints executable, compressing understanding-side structure into an ordered hidden action trajectory before visual synthesis.

\subsection{Role and Computation Ablation}
\label{sec:ablation}

\begin{wrapfigure}{r}{0.41\textwidth}
\vspace{-0.8em}
\centering
\includegraphics[width=\linewidth]{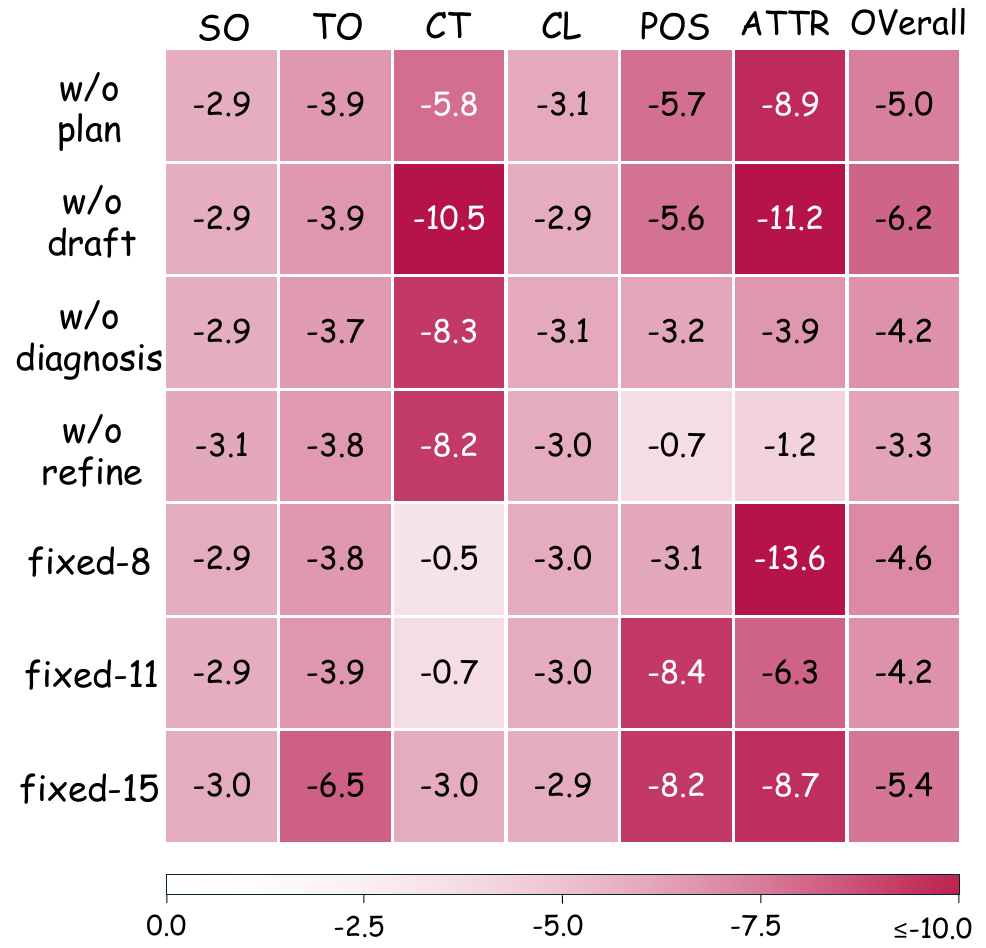}
\caption{
GenEval drops under role removals and fixed latent budgets.
}
\label{fig:LAC-ablation}
\vspace{-1.0em}
\end{wrapfigure}

We next examine which roles and computation budgets drive compositional control. Figure~\ref{fig:LAC-ablation} reports category-wise GenEval drops from full \model{} under role removals and fixed-length variants, with backbone, data, and optimization held fixed.

The role ablations show that the four roles provide complementary control. Removing \texttt{draft} causes the largest overall drop, especially on Counting and Color Attribute, indicating that the internal visual hypothesis is important for object-slot tracking and attribute binding. Removing \texttt{plan} mainly hurts Position and Color Attribute, suggesting that global intent and relational setup help constrain synthesis. \texttt{diagnosis} and \texttt{refine} have smaller overall effects but remain important for Counting, consistent with their role in correcting missing or mismatched objects.

The fixed-length variants show that more latent computation is not always better. Fixed-8 strongly degrades Color Attribute, while Fixed-15 still underperforms on Position and Color Attribute, suggesting that both insufficient and excessive latent budgets can weaken control. This supports adaptive role-wise halting, where \model{} allocates latent computation according to the prompt rather than using a single fixed budget.

\subsection{Qualitative Evaluation}
\label{sec:qualitative}

Appendix~\ref{app:qualitative} provides qualitative comparisons and additional visual samples. On reasoning-intensive prompts, BAGEL often captures the coarse scene but misses fine-grained constraints such as spatial relations, attribute bindings, material properties, object order, and knowledge-specific cues. \model{} better preserves these constraints, suggesting that the learned action trajectory helps maintain prompt-relevant structure during synthesis. These examples align with the quantitative results on GenEval and WISE, where the largest gains appear on relation-, binding-, and knowledge-sensitive categories.
\section{Conclusion}

We introduced \model, a framework that makes reasoning actionable for unified image generation by converting inferred constraints into hidden continuous actions. Instead of exposing reasoning as text or using it only after generation, \model rolls out a role-structured latent trajectory and writes it into the hidden stream as generation-time control, trained with prior-guided variational alignment, draft-image grounding, supervised halting, and LF-GRPO. Experiments show consistent gains on compositional and knowledge-grounded generation, while ablations and latent interventions confirm that the learned actions are actively consumed by the generator rather than serving as unused latent capacity. These results suggest that unified image generation benefits when understanding is converted into executable hidden controls, and future work should reduce teacher dependence, improve RL efficiency, and extend latent action control to broader unified backbones.

\newpage
{
\small
\bibliographystyle{plainnat}
\bibliography{refs}
}

\end{document}